\documentclass[sigconf,9pt]{acmart}

\usepackage{algorithmic}
\usepackage[ruled,vlined,linesnumbered]{algorithm2e}
\SetAlgoSkip{}
\usepackage{graphicx}
\usepackage{subcaption} 

\AtBeginDocument{%
  }

\copyrightyear{2024}
\acmYear{2024}
\setcopyright{acmlicensed}\acmConference[DAC '24]{61st ACM/IEEE Design
Automation Conference}{June 23--27, 2024}{San Francisco, CA, USA}
\acmBooktitle{61st ACM/IEEE Design Automation Conference (DAC '24), June
23--27, 2024, San Francisco, CA, USA}
\acmDOI{10.1145/3649329.3658491}
\acmISBN{979-8-4007-0601-1/24/06}

\newcommand{\conclave}{\textit{CONClave}}  % don't add a space to this 

\thanks{This work was partially supported by funding from National Science Foundation grants CPS 1645578 and Semiconductor Research Corporation (SRC) project 3154.}

\begin{document}

\title{\conclave{} - Secure and Robust Cooperative Perception for CAVs Using Authenticated Consensus and Trust Scoring}


\author{Edward Andert}
\affiliation{%
  \institution{Arizona State University}
  \city{Tempe}
  \state{Arizona}
  \country{USA}}
\email{edward.andert@asu.edu}

\author{Francis Mendoza}
\affiliation{%
  \institution{Arizona State University}
  \city{Tempe}
  \state{Arizona}
  \country{USA}}
\email{fmendoz7@asu.edu}

\author{Hans Walter Behrens}
\affiliation{%
  \institution{Arizona State University}
  \city{Tempe}
  \state{Arizona}
  \country{USA}}
\email{hwbehren@asu.edu}

\author{Aviral Shrivastava}
\affiliation{%
  \institution{Arizona State University}
  \city{Tempe}
  \state{Arizona}
  \country{USA}}
\email{aviral.shrivastava@asu.edu}

\renewcommand{\shortauthors}{Andert et al.}

\begin{abstract}
Connected Autonomous Vehicles have great potential to improve automobile safety and traffic flow, especially in cooperative applications where perception data is shared between vehicles. However, this cooperation must be secured from malicious intent and unintentional errors that could cause accidents. Previous works typically address singular security or reliability issues for cooperative driving in specific scenarios rather than the set of errors together. In this paper, we propose \conclave{} -- a tightly coupled authentication, consensus, and trust scoring mechanism that provides comprehensive security and reliability for cooperative perception in autonomous vehicles. \conclave{} benefits from the pipelined nature of the steps such that faults can be detected significantly faster and with less compute.  Overall, \conclave{} shows huge promise in preventing security flaws, detecting even relatively minor sensing faults, and increasing the robustness and accuracy of cooperative perception in CAVs while adding minimal overhead.
\end{abstract}

\keywords{}


\maketitle

\section{Introduction}
Cooperative autonomous vehicle operation has the potential to make roadways dramatically more safe and efficient \cite{khayatian2020survey}. 
Even if all the technicalities of executing a cooperative maneuver can be solved, there are still problems securing them from unauthorized participants. Beyond that, there is the issue of preventing both malicious vehicles that intentionally disrupt a cooperative application as well as preventing faulty vehicles that unintentionally disrupt a cooperative application due to an error. For example, a malicious vehicle could try to get ahead in the queue for a cooperative intersection by falsifying data. A fault, on the other hand, could be an autonomous vehicle that has a sensor malfunction and is sharing bad data with another vehicle that is blindly trusting the data to see around a corner that is out of its sensor range. Even if the exact reaction to these types of situations could be vehicle or vendor specific, the overarching identification and prevention of all disrupting vehicles, whether intentional or not, is paramount to running successfully cooperation amongst autonomous vehicles.

Detecting malicious and unintentional faults requires multiple steps, including authentication and verification of incoming data \cite{lu2018survey}. However, without a trusted third party with its own sensors involved in every vehicle area network, the scope increases to include consensus \cite{he2019cooperative}. Existing state of the art methods typically treat authentication, consensus, and trust scoring either separately or together in a limited subset of cooperative scenarios. Guo \textit{et al.} propose a method to log events using blockchain, but their approach completely ignores the problem of keeping out unauthorized participants and also does not have any mechanism to keep authenticated users from making up events \cite{guo2020proof}. More recently, trust scoring methods have been used in place of proof of work. For instance, Mankodiya \textit{ et al.} \cite{mankodiya2021xai} use a specialized ML bases trust scoring that could take the place of proof of work but it is not coupled with a consensus method and therefore cannot reap those extra benefits. Bhattacharya \textit{et al.} do tackle the authentication and consensus problems at once, but too many assumptions are made for the specific application, and therefore their approach will not work for general cooperative scenarios \cite{bhattacharya20226blocks}. 

This paper presents \conclave{} -- an application-level network protocol designed for sensor networks that require reliable and trustworthy data in the context of Cooperative Autonomous Vehicles (CAVs) and Cooperative Infrastructure Sensors (CISs). The three primary contributions of \conclave{} are:
\begin{enumerate}
  \item A three party homomorphic hashing based authentication process which includes the manufacturer, a third party authority/government, and the vehicle itself. This inclusion ensures that all entities (CAVs and CISs) that wish to participate in the system must have the approval of both the manufacturer and governmental stakeholders.
  \item A BOSCO-based single-shot consensus protocol that works in a dynamically changing geo-spatial vehicular networks by limiting the latency and resource requirement of the consensus protocol on non-discrete sensed values. Instead of generating consensus on a common world-view, \conclave{} generates consensus on the individual world-view provided by each agent. This eliminates Byzantine attacks on the network, leaving the common world-view generation work for the next sensor fusion step.
  \item A perception trust scoring technique that reports an accuracy score by utilizing sensor and recognition pipeline characterization data as the accuracy predictor, allowing for errors to be detected down to the individual sensor level that are not picked up by other state of the art methods. This trust scoring technique is tightly coupled with the authentication and consensus step so that it can operate in place of a proof of work to improve real-time performance.
\end{enumerate}

\conclave{} was tested against the state of the art trust scoring method Trupercept \cite{hurl2020trupercept} using fault and malicious injection on a 1/10 scale model autonomous vehicles using a motion capture system as ground truth. 1100 faults and malicious attacks were injected over the course of 14 different scenarios while varying the severity and number of the fault/injection. \conclave{} detected 96.7\% of the 300 sensor extrinsic faults injected, 83.5\% of the 300 software faults injected, 67.3\% of the 300 malicious injections and removals, and 100\% of the 200 communication faults and malicious injections that we subjected it to. On the other hand, the state of the art method TruPercept only detected 29.6\% of sensor extrinsic faults, 34\% of software faults, 32.6\% of malicious injections and removals, and 19.6\% of the communication faults and malicious injections. Overall, \conclave{} had a mean time to detection that was 1.83x faster on average and 6.23x faster in the best case when compared to TruPercept on the faults TruPercept could detect. 
%while being able to be run in real time on embedded hardware.

\section{Related Work}
\noindent\textbf{Authentication:} When distributed agents communicate in the field, authentication is critical or the network is open to Sybil attacks \cite{lu2018survey, dibaei2020attacks}. Further complicating the issue, conditions often prevent real-time communication with a central server, and local resource constraints limit processing and storage \cite{dharminder2021edge}. Handy \textit{et al.} assume a more difficult task with no centralized authority or setup phase, but participants establish keys with each new participant through a process that does not consider the Sybil threat \cite{nandy2021secure}. Wang \textit{et al.} rely on specialized hardware such as Physically Uncloneable Functions (PUFs), an impractical choice for real-world deployments \cite{wang2021blockchain}. Similarly, approaches that rely on trusted execution environments (TEEs) are susceptible to eventual compromise and key extraction (e.g. via cold boot attacks \cite{halderman2009lest} or side channels \cite{liu2022frequency}). To address these challenges, we use a three-way knowledge partitioning between a government entity, manufacturer, and each individual participant. To allow for reconstruction, we rely on an approach that allows for intermediate hash composition using the homomorphic hash tree described by Behrens \textit{et al.} \cite{behrens2020pando}. Though this produces larger hashes, it allows for asymmetric reassembly of hashes, compartmentalizing information, and preventing the compromise of any one party from undermining the security of the authentication protocol \cite{bellare1997new}.

\noindent\textbf{Consensus:} In a distributed environment, cooperative perception algorithms can quickly succumb to byzantine faults \cite{dibaei2020attacks}. Whether due to communication dropout or malicious intent, faults will manifest themselves as data corruption in the subsequent sensor fusion step. A popular way to solve these issues is byzantine fault tolerant consensus, however, consensus on non binary values is slow. Han \textit{et al.} solve this by eschewing the need for consensus on all sensor values by bounding the problem to just nearby vehicle positions in a platoon in addition to many other specializations \cite{han2022distributed}. However, this approach will not work for general cooperative perception. To address this challenge, our approach relies on a semi synchronous distributed Byzantine tolerant consensus on the data each party sent, rather than coming to consensus on the correctness of that data. The correctness proof is left for the subsequent trust scoring step in the pipeline. This technique keeps the consensus itself lightweight and eschews the need for any proof of work by using the trust score of the sensed values computed next as substitute.

\noindent\textbf{Trust Scoring:} In a cooperative perception environment, a minor disagreement in sensor input caused by a sensor fault or malicious actor could result a catastrophic incident and must be prevented \cite{khayatian2020survey, hbaieb2022survey}. Cavorsi \textit{et al.} propose a method to apply a trust score against robots sensing local traffic in their region that can detect adversaries and lower the percentage error in the locally fused traffic estimate \cite{cavorsi2022exploiting}. However, it is not clear how this method can be generally applied to cooperative perception nor does it take into account the expected accuracy of the sensors involved. Hurl \textit{et al.} propose a cooperative perception specific trust scoring method and test it using simulated data \cite{hurl2020trupercept}. The trust score is applied to a sensor fusion algorithm as a weight, and the result is a better sensor fusion. Their method is limited to the case that the CAV sensor configurations are uniform, containing both a camera and a LIDAR as they use the camera confidence as the expected accuracy of each sensed object and the LIDAR point count as a proxy for the visibility. To address this, we create a trust scoring system that uses a generalized error estimation technique for heterogeneous sensing platforms borrowed from Andert \textit{et al.} while consuming the results of the previous consensus step to prevent byzantine faults \cite{andert2022paramertized}.

\section{Our Approach}
\subsection{Overview}
\vspace{-10pt}

\begin{figure}[h]
  \centering
    \includegraphics[width=.4\textwidth]{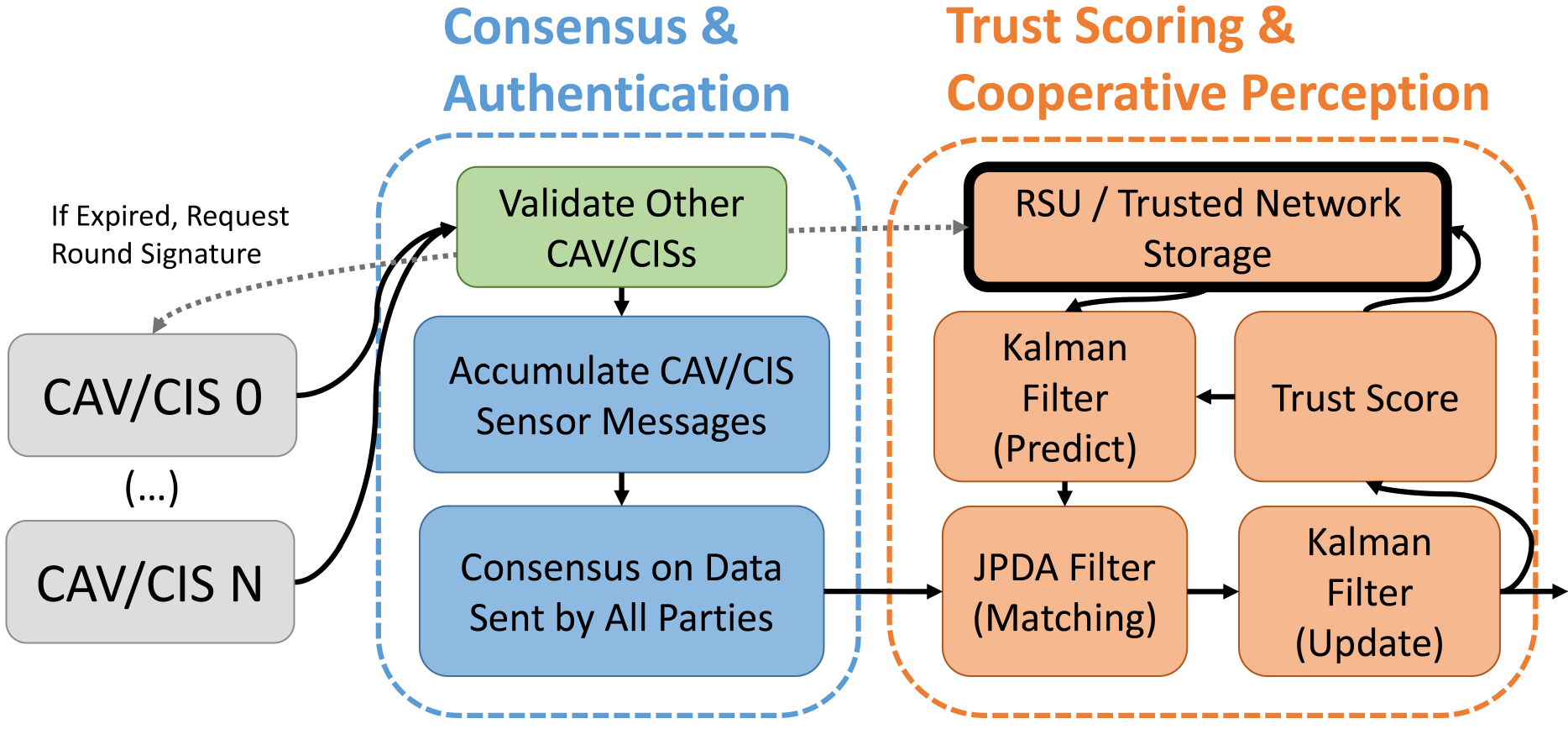}
    \vspace{-8pt}
    \caption{Overview of \conclave{}. Consensus and Authentication steps occur concurrently to reach a sensor data set that all participant CAVs and CISs agree upon. The sensor data set is then taken as input to our cooperative perception and trust scoring steps resulting in trust scores for each participant.}
    \label{fig:overview}
    \vspace{-10pt}
\end{figure}

To achieve a secure consensus and trust scoring of CAVs and other sensing infrastructure for reliable cooperative driving, \conclave{} proposes a three-step process, which can be seen depicted at a high level in figure \ref{fig:overview}. First, all participants are authenticated to address the risk of Sybil attacks. We create a novel authentication scheme that leverages homomorphic hashing, incorporates both the manufacturer and a government entity, and allows participants to authenticate each other in such a way that participants in a consensus round don't always need to have communication with a trusted RSU. Next, we come to consensus on the sensor values that all participants submit to the consensus round using Byzantine fault tolerant consensus protocol such that faults in communication can such as packet delay or dropped messages don't manifest themselves later as error in the output \cite{sabahi2011security}. We come to consensus on the sensor values that each participant sends such that we reduce the computation time by bounding the problem to be consensus on the sensor values each participant sent using the Bosco consensus protocol, which was modified to be semi-synchronous \cite{song2008bosco}. Finally, a trust scoring technique is applied to the sensing input set that results from the consensus round, to verify the correctness of the data each participant sent. Instead of using camera confidence values as an accuracy indicator like Hurl \textit{et al.} use, we use parameterized sensor pipeline accuracy values from Andert \textit{et al.} \cite{hurl2020trupercept, andert2022paramertized}. This, along with being closely coupled with a sensor fusion technique, allows our trust scoring to be both fast and more accurate than the previous state of the art. Our trust scoring not only improves the accuracy of cooperative perception, it also serves as a replacement of the proof of work for our consensus step. All of this combines to prevent most known attack vectors and errors that can occur in a cooperative perception environment. Next, we explain the three steps of \conclave{} and its working in more detail.

% \vspace{-5pt}
\subsection{Three Party Authentication}
% \vspace{-10pt}

\begin{algorithm}
\caption{Three party authentication setup.}
\label{alg:authentication}
\KwData{ego, mnf, gov, nearbyRSU, expirationTime}
\uIf{ego->keyPair == false}{
  ego.$P_c$, ego.$S_c$, ego.$UUID_c$ = genKeyPair()\;
  ego.$\textit{Chal}_c$ = gov.genChal(ego.$UUID_c$, mnf.$S_m$, gov.$S_g$)\;
  ego.$\textit{Resp}_c$ = gov.genResp(ego.$UUID_c$, mnf.$S_m$, gov.$S_g$)\;
  ego.keyPair = true\;
}
\uIf{ego.roundToken.age() >= .9 * expirationTime}{
    \uIf{nearbyRSU.withinRange(ego.position)}{
        \uIf{roundNum.age() > expirationTime}{
            roundNum++\;
            $UUID_r$ = genUUID()\;
            $sig_r$ = genRoundSig(roundNum, $UUID_r$, mnf.$S_m$, gov.$S_g$)\;
            transmitAllRSUsSecure(roundNum, $sig_r$, $UUID_r$)\;
            ego.$sig_r$ = transmitAllEgosSecure($sig_r$)\;
        }
        ego.$\textit{Chal}_t$ = nearbyRSU.genChal(ego.$UUID_c$, ego.$\textit{Chal}_c$, $sig_r$)\;
        ego.$\textit{Resp}_t$ = nearbyRSU.genResp(ego.$UUID_c$, ego.$\textit{Chal}_c$, $sig_r$)\;
    }
}
\end{algorithm} 

A high level depiction of our authentication setup process can be seen in algorithm \ref{alg:authentication}. To initialize, both the law enforcement/governmental agency and manufacturer generate a secret key known only to themselves denoted as $S_g$ and $S_m$ respectively. The manufacturer generates an asymmetric keypair $P_c, S_c$ and stores it locally on the vehicle; crucially, we do not require any central database of these keys (line 2). In an interactive process, the two players exchange the CAV/CIS's identity and generate a challenge $\textit{Chal}_c$ and response hash $\textit{Resp}_c$ which is also stored locally (line 3-5). This inclusion ensures that all CAVs/CISs which wish to participate in the system have the approval of both stakeholders. Note that neither party exchanges their secret keys $S_m$ or $S_g$, instead using hashed versions $\textit{Chal}_c$ and $\textit{Resp}_c$ to prevent inappropriate use.

Next, both stakeholders must periodically go through an interactive process to refresh what we call as a \textit{round signature} $\textit{Sig}_r$ -- a validity mechanic that allows for either party to exit from participation (lines 8-13). Stakeholders may tune the frequency of this process to increase or decrease the duration in which CAVs/CISs may operate asynchronously. Once generated, these signatures $\textit{Sig}_r$ are securely distributed to each RSU. As vehicles travel within range of an RSU, they may choose to issue a renewal request to that RSU (line 7). These messages are encrypted with a CAV/CIS's private key, and the corresponding public key is transmitted along with the request to provide authenticity but not secrecy. Nonces prevent replay attacks. The RSU may optionally check the CAV/CIS's identifier against a central database to ensure compliance, such as valid licensing or inspection requirements. Once the RSU validates the request, ephemeral challenge and response \textit{tokens} $\textit{Chal}_t$ and $\textit{Resp}_t$ are generated and encrypted with the CAV/CIS's public key before sending them back (lines 14, 15). This ensures that eavesdroppers may not re-use a CAV/CIS's token, as they lack the corresponding private key. Depending on how often rounds change, it may be desirable to store the subsequent tokens to ensure that validation can take place between CAVs/CISs whose round tokens differ in sequence by one. 

% \vspace{-5pt}
\subsection{Single-shot Consensus}

For consensus rounds, we utilize a set area around an intersection with a constant trigger. All CAVs/CISs within range attempt to participate in the round and local IPs are known ahead of time. When establishing a relationship between CAVs/CISs during a given consensus step, each CAV/CIS provides additional metadata with their broadcast to allow for authentication. Each CAV/CIS generates this metadata $\textit{Chal}_{c1}$, and shares it along with a hashed version of its ID $ID_{c1}$ and its public key $P_{c1}$. Recipients check the received values using their own local tokens $ID_{c2}$ and $\textit{Resp}_{t2}$ to ensure compliance, and if valid, they temporarily store the public keys to allow for secure communication during the consensus step.

Next, participants accumulate the sensing messages from all other participants. This stops when a message is received from every known participant or the sensing transmission timeout is reached. Each participant sends out the accumulated set of sensing messages, received with valid authentication, and accumulates the same message from other participants. This stops when a message is received from every known participant or the aggregate transmission timeout is reached. Finally, each participant decides their vote according to the BOSCO algorithm and sends the result to all other participants, as well as nearby trusted RSUs for secure storage \cite{song2008bosco}.

% \vspace{-5pt}
\subsection{Accurate Trust Scoring}

\begin{algorithm}
\caption{Trust Scoring.}
\label{alg:trust_scoring}
\KwData{trustScores, sensorData, tracks, participants}
\KwResult{trustScores, tracks}
tracks.predictEKF()\;
tracks.JPDAFAssociation(sensorData)\;
prelimResult = tracks.updateUKF(sensorData, trustScores)\;
exists = tallyExistenceVotes(sensorData, participants)\;
sensorDataTrun = remNonByzantine(exists, sensorData)\;
trustScores = calcSSDS(prelimResult, sensorDataTrun)\;
trustScores = enforceMinimums(trustScores)\;
tracks = tracks.updateUKF(sensorDataTrun, trustScores)\;
\end{algorithm}

Our trust scoring method is closely coupled with a UKF based sensor fusion method, depicted in algorithm \ref{alg:trust_scoring}. It can be ran by all participants separately or alternatively ran on a nearby trusted RSU and distributed out. The first step of this is matching observations to tracks using a JPDA Filter while taking into account expected error and bounding box size \cite{svensson2011set}. We utilize the Unscented Kalman Filter (UKF) approach that Andert \textit{et al.} use for fusion (lines 1-3) \cite{andert2022paramertized}. Utilization of observations that are coming from each vehicle is conditional upon the existing trust score, or sensing standard deviation score ($SDS$). If $SDS$ exceeds a certain value $SDS_{max}$, the sensor platform is considered not trustworthy and none of its observations nor its own position will be included in this global fusion.

\conclave{} looks at two factors when calculating the trust score: i) Was an object supposed to be detected or not?, and ii) Was an object detected with the accuracy it was expected to be detected with? In order to evaluate the first, we determine if an object should have been seen or not with respect to other sensor platforms using a Byzantine tolerant voting scheme (line 4,5). We iterate through all the tracks and mark the track as existing if the object is within the FOV and range of a sensor as well as the visibility percentage threshold. Our method maintains byzantine tolerance by requiring that a majority of all participants should see an object according to their FOV and modeled obstructions to vote whether a track exists or not. For the second item, we utilize the estimated accuracy of the local fusion output of each vehicle as defined by Andert \textit{et al.} \cite{andert2022paramertized}, to determine the expected accuracy of each detection.

All participants within the sensor network need to have a minimum accuracy boundary, otherwise a possible attack vector would be to report its sensors as incredibly inaccurate. For sensing pipelines, we enforce a minimum sensing range requirement, a minimum FOV requirement, and a minimum sensor error magnitude requirement within this range and FOV. For localization pipelines, we enforce a minimum error magnitude requirement. If any participant's sensed values exceed these thresholds, the participant values will not be considered in the trust scoring and should be sent for repairs, even though it passes the authentication and consensus rounds.

Using the matching results from the JPDA filter and bounding box step as well as the fused positions outputted by the UKF for observability consensus, we have the necessary ingredients to perform a trust scoring of the reported accuracy of the sensor platform versus the fused position from the consensus round (line 6). If a track is reported as \textit{exists}, the reported accuracy of the detected value of the CAV / CIS is compared with that contained in the fused output of the UKF. For the sensed values from each CAV/CIS that were matched to the track, the reported position $<x,y>$ (or $z_k$) is subtracted from the estimated position produced by the UKF (or $\hat{x}_{k|k}$), shown in the numerator of equation \ref{equation:sds}. ${P}_{k|k}$ returned by the UKF encompasses the expected error of the measurements from all sensors involved in the local fusion as well as the estimation from the UKF itself \cite{andert2022paramertized}. We then normalize by the expected error $\mathbb{E}(\mu^{\alpha}_{\theta})$ which is contained in $\Sigma^{\alpha}_{\theta}$ and can be extracted using the eigenvalues, shown in the denominator of equation \ref{equation:sds}. The result is why this is called the standard deviation score ($SDS$) as this is simply the standard deviation of the measured error returned by the global fusion \textit{w.r.t} the value of the expected error from the measurement covariance in the local fusion of the sensor platform. %Before this $SDS$ value is used, there are some checks that need to be passed first.

In order to keep a sensor from reporting itself as accurate in a vacuum, we enforce a rule of three -- meaning at least three sensors must be matched and detecting the same track for a standard deviation frame to be added to the SDS revolving buffer for that object (line 7). This is in addition to the byzantine tolerant consensus on the existence or lack of existence of the track from all sensors. Furthermore, to keep low confidence tracks from being levied against a sensor, we enforce the second piece of the rule of three which dictates that the hypotenuse of the accuracy reported by the $roundFusion$ track (or ${P}_{k|k}$) must be three times as accurate as the hypotenuse of the accuracy reported by the sensor itself (or $\Sigma^{\alpha}_{\theta}$). If all of these conditions hold true, then it is prudent that the sensor $\alpha$ is attributed with the $SDS$ by placing it in the last position of the revolving buffer, which is averaged to create the overall trust score. The rule of three applies to missed detection, three sensors must agree the object is there along with the consensus that the track exists. If these constraints are met, any sensor platform that should detect that object will have a missed detection frame added. The value $\rho$ is then added in place of the $SDS$ frame for the sensor platform that did not detect the track. $\rho$ was set to $3 * min\_Sensor\_Accuracy$. 

\vspace{-10pt}
\begin{equation}
SDS^{\alpha}_\theta = \frac{\sqrt{(\lambda^{\downarrow}_{0}(\mu^{\alpha}_{\theta}) - \lambda^{\downarrow}_{0}(\hat{x}_{k|k_{\theta}}))^2 + (\lambda^{\downarrow}_{1}(\mu^{\alpha}_{\theta}) - \lambda^{\downarrow}_{1}(\hat{x}_{k|k_{\theta}}))^2}}{\sqrt{(\lambda^{\downarrow}_{0}(\Sigma^{\alpha}_{\theta}) - \lambda^{\downarrow}_{0}(P_{k|k_{\theta}}))^2 + (\lambda^{\downarrow}_{1}(\Sigma^{\alpha}_{\theta}) - \lambda^{\downarrow}_{1}(P_{k|k_{\theta}}))^2}}
\label{equation:sds}
\end{equation}

% \begin{equation}
% SSDS_\alpha = \sum_{i=0}^{l} SSDSBuffer^{\alpha} 
% \label{equation:ssds}
% \end{equation}

For each participant in the round, a new SDS score is calculated in equation \ref{equation:sds}. This score is then added to the $SSDS$ buffer (line 6). The set of $SSDS$ scored for each participant constitutes their trust score. This value is shared globally among RSUs and is updated after each consensus round ends. Finally, the tracks are updated using the curated sensor set and new trust scores (line 8).

\section{Experimental Setup}

% \usepackage{graphicx}
\begin{table}[]
\resizebox{\columnwidth}{!}{%
\begin{tabular}{|l|l|}
\hline
\textbf{Error\#: Error Name}       & \textbf{Description}                                                                                                                      \\ \hline
E1: Camera Shift       & \begin{tabular}[c]{@{}l@{}}For CAV I, Camera extrinisics skewed N degrees.\end{tabular}                                          \\ \hline
E2: LIDAR Shift        & \begin{tabular}[c]{@{}l@{}}For CAV I, LIDAR extrinisics skewed N degrees.\end{tabular}                                           \\ \hline
E3: Cam \& LIDAR Shift & \begin{tabular}[c]{@{}l@{}}For CAV I, Camera and LIDAR extrinsics\\ are skewed by the same N degrees.\end{tabular}                       \\ \hline
E4: Random Data Loss          & \begin{tabular}[c]{@{}l@{}}For CAV I, with probability N, each\\ detection the vehicle has may be removed.\end{tabular}             \\ \hline
E5: Malicious Removal     & \begin{tabular}[c]{@{}l@{}}For CAV I, with probability N, the CAV\\ removes vehicles crossing the intersection.\end{tabular}       \\ \hline
E6: Malicious Insertion   & \begin{tabular}[c]{@{}l@{}}For CAV I, with probability N, the CAV\\ injects vehicle detections into the intersection.\end{tabular} \\ \hline
E7: Localization            & \begin{tabular}[c]{@{}l@{}}For CAV I, the localization error that CAV I\\ is experiencing will be increased by N percent.\end{tabular}    \\ \hline
E8: Local Sensor Fusion     & \begin{tabular}[c]{@{}l@{}}For CAV I, at the local fusion level the \\ covariance of the LIDAR is decreased by N\%.\end{tabular}          \\ \hline
E9: Global Sensor Fusion    & \begin{tabular}[c]{@{}l@{}}For CAV I, at the global fusion level the\\ covariance of the CAV I is decreased by N\%.\end{tabular}          \\ \hline
E10: Unauthorized User            & CAV I has an invalid authentication challenge.                                                                                            \\ \hline
E11: Expired Round Token      & CAV I has a round token that has expired.                                                                                                     \\ \hline
E12: Byzantine Fault              & \begin{tabular}[c]{@{}l@{}}CAV I drops a packet when sending a message.\end{tabular}                             \\ \hline
E13: Replay Attack                & \begin{tabular}[c]{@{}l@{}}CAV I replays data another participant\\ sent one round before.\end{tabular}                                   \\ \hline
E14: Spoofed Localization         & \begin{tabular}[c]{@{}l@{}}CAV I sends the wrong location for itself.\end{tabular}                                  \\ \hline
\end{tabular}%
}
\label{tab:error_descriptions}
\caption{Descriptions and examples of the 14 error injection tests we performed against \conclave{} to test resilience.}
\vspace{-20pt}
\end{table}

\begin{figure}[ht]
  \centering
    \includegraphics[width=.35\textwidth,height=2.5cm]{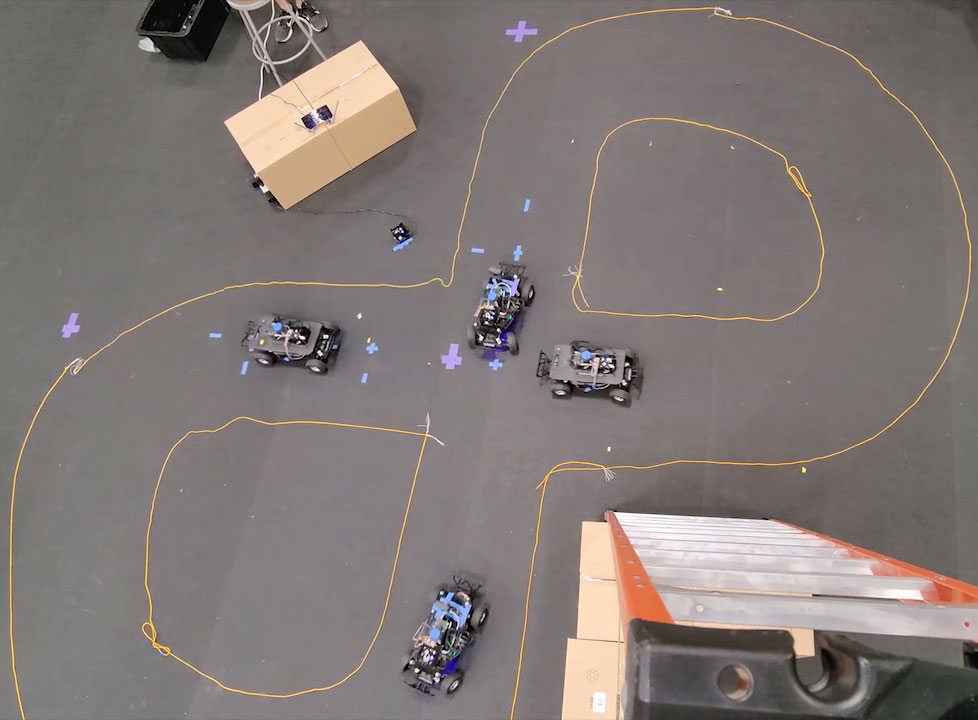}
    \caption{Four one-tenth scale CAVs with IMX160 camera, Slamware M1M1 LIDAR, and Nvidia Jetson Nano for on-board processing along with two one-tenth scale CIS traffic cameras using Jetson Nano and IMX160 camera test setup.}
    \vspace{-10pt}
    \label{fig:one_tenth_cam}
\end{figure}

For testing, we utilize 1/10 scale autonomous vehicle replicas. Our setup consists of four scale CAVs with a front facing 160 degree FOV camera and a 360 degree FOV single channel LIDAR and two mounted CISs with a 160 degree FOV camera. Figure \ref{fig:one_tenth_cam} shows our setup with four CAVs.

Using data collected from a set of 10, ten-minute-long tests for each physical configuration, we perform error injection with the 14 scenarios shown in Table 1. The simplest scenarios are sensor errors that can be easily caused in any autonomous vehicle by jarring a sensor. For E1-E3, the same data from the sensors are used, but the extrinsics of the sensors will be skewed by $N$ degrees resulting in a shift in the data from that sensor. The next category of error is malicious error in which we purposely inject or remove detection with probability $N$. Next, we have software errors that manifest itself as a bad weighting in the sensor fusion where we change weightings by some $N$ percent. Finally, we have communication faults and attacks where we cause $N$ vehicles in the simulation to experience a communication error. The tests are run for a random amount of time from 120 seconds through 540 seconds with normal operation before we begin injecting the specific fault. If the trust score for a vehicle becomes 1.2x of the baseline within 60 seconds after the fault is injected, we consider the fault to be caught and record the MTTD. Each fault injection was run 10 times at each step for a total of 1100 tests.

\vspace{-5pt}
\vspace{-5pt}

\section{Results}
% \begin{figure*}[ht]
%   \centering
%     \includegraphics[width=\textwidth]{images/conclave_errors.png}
%     \caption{Top Left: Mean time to detection (MTTD) for sensor extrinsics errors. \conclave{} detects as little as a two degree camera shift (E1) or one degree shift in the LIDAR (E2, E3) while TruPercept fails to detect all but E3. Bottom Left: MTTD of malicious and accidental injections and removals of tracks. \conclave{} detects most injection/removals on par or better than TruPercept. Top Right: Mean time to detection (MTTD) of localization errors. TruPercept fails to detect most of these while Conlcave succeeds. Bottom Right: Other attacks and faults with one or two participants. \conclave{} detects almost all of these within the first frame fast while TruPercept fails at all except localization spoofing. Note E10-E13 are not plotted for TruPercept as it does not detect them.}
%     \label{fig:mttd}
% \end{figure*}

\begin{figure*}
  \centering
  \begin{subfigure}{0.49\textwidth}
    \vspace{-10pt}
    \includegraphics[width=\textwidth]{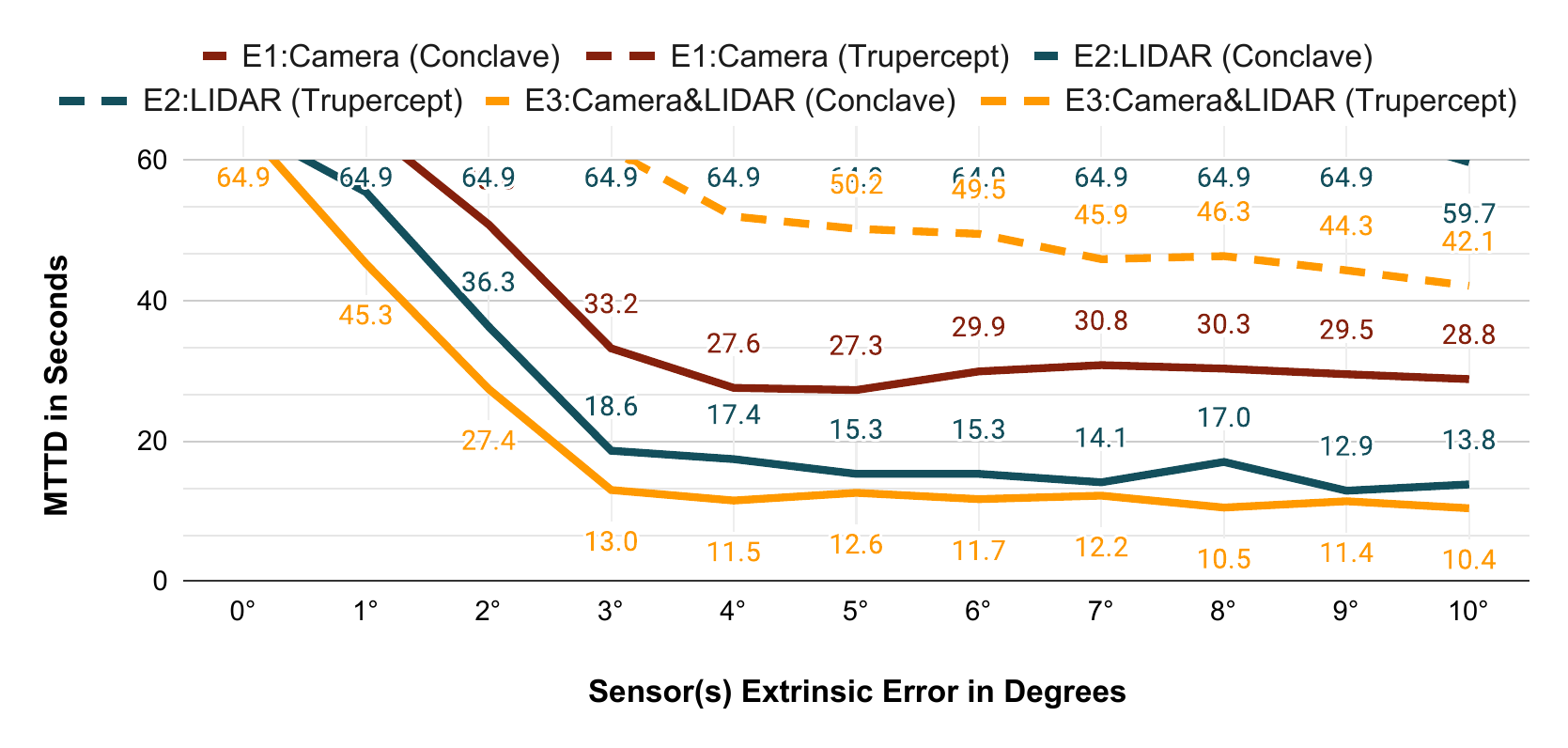}
    \vspace{-25pt}
    \caption{Mean time to detection (MTTD) for sensor extrinsics errors. \conclave{} detects as little as a two degree camera shift (E1) or one degree shift in the LIDAR (E2, E3) while TruPercept fails to detect all but E3.}
    \label{fig:e123_mttd}
    \vspace{-10pt}
  \end{subfigure}
  \hfill
  \begin{subfigure}{0.49\textwidth}
    \vspace{-10pt}
    \includegraphics[width=\textwidth]{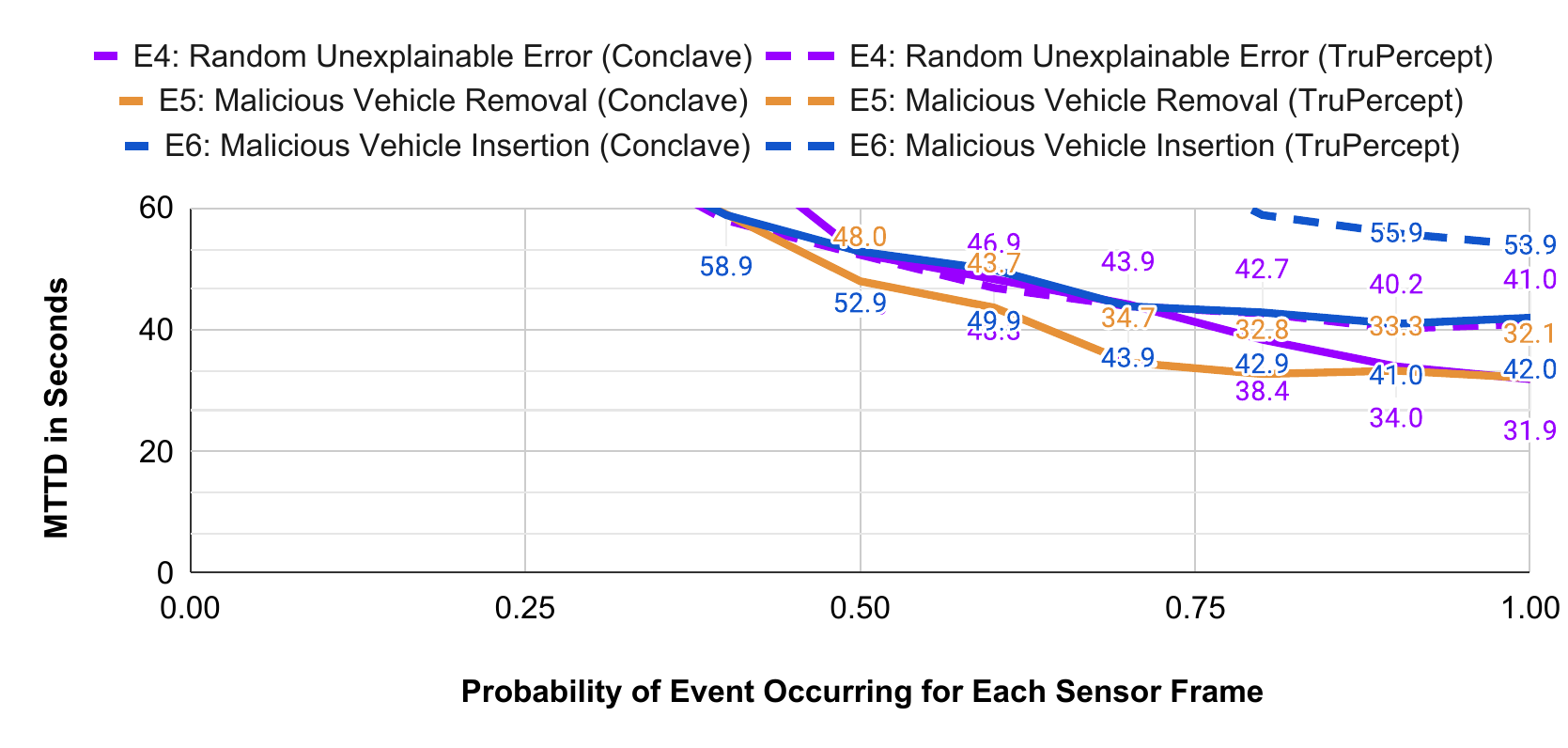}
    \vspace{-25pt}
    \caption{MTTD of malicious and accidental injection/removals of tracks. \conclave{} detects most injection/removals on par or better than TruPercept.}
    \label{fig:e456_mttd}
    \vspace{-6pt}
  \end{subfigure}
  \medskip
  \begin{subfigure}{0.49\textwidth}
    \includegraphics[width=\textwidth]{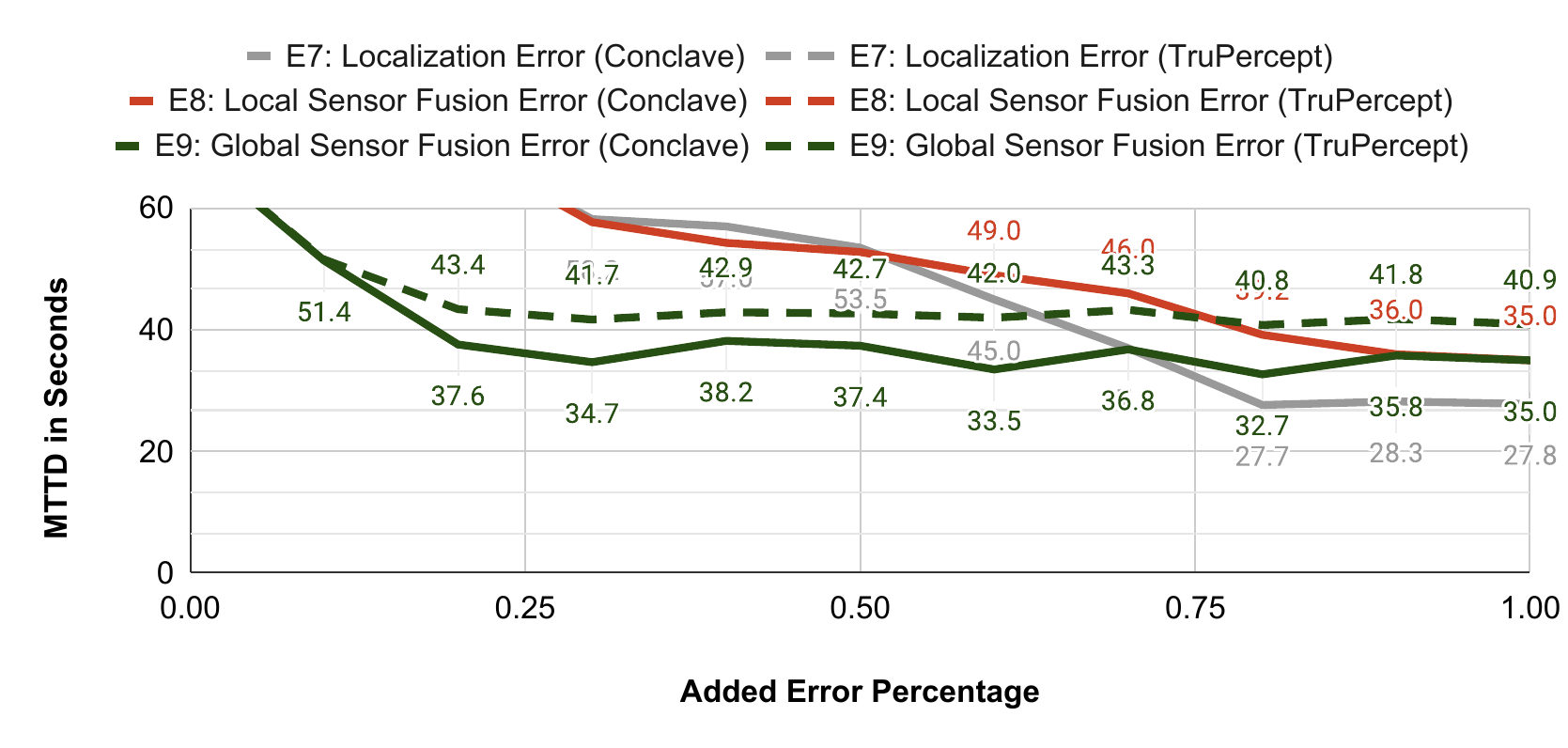}
    \vspace{-25pt}
    \caption{Mean time to detection (MTTD) of localization errors. TruPercept fails to detect most of these while Conlcave succeeds.}
    \label{fig:e789_mttd}
    \vspace{-10pt}
  \end{subfigure}
  \hfill
  \begin{subfigure}{0.49\textwidth}
    \includegraphics[width=\textwidth]{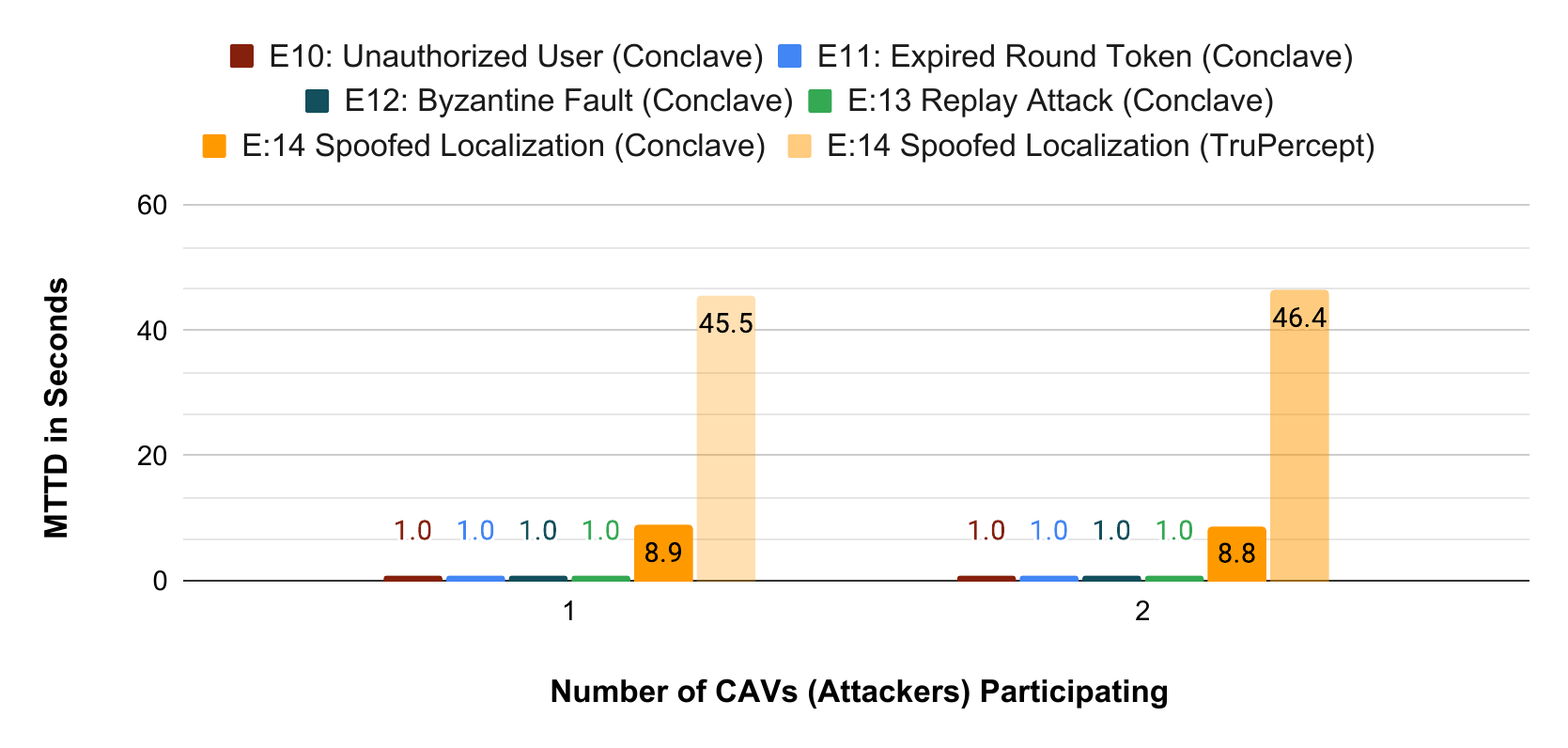}
    \vspace{-25pt}
    \caption{\conclave{} detects almost all of these within the first frame fast while TruPercept fails at all except localization spoofing. Note E10-E13 are not plotted for TruPercept as it does not detect them.}
    \label{fig:eothers_mttd}
    \vspace{-10pt}
  \end{subfigure}
  \label{fig:mttd}
\end{figure*}

\noindent{\textbf{5.1 - \conclave{} quickly detects nearly all extrinsics errors}}
\newline E1-E3 where the sensors experience a physical sensor shift of N degrees are shown in figure \ref{fig:e123_mttd}. These tests showcase \conclave{}'s ability to pick out relatively minor errors within the cooperative perception environment. Errors as minor as a two-degree camera shift, one-degree LIDAR shift, or one-degree shift combination are caught. \conclave{} detects an impressive 96.7\% of the 300 tests. TruPercept is only able to catch large magnitude errors, such as a ten-degree LIDAR shift or a four-degree or more camera and LIDAR shift resulting in a detection rate of 29.6\% of the 300 tests. TruPercept does not have a suitable predictor that works when the IOU match is still high, but the error variance is higher than it should be, instead relying on the confidence of the camera as a predictor \cite{hurl2020trupercept}. Therefore, TruPercept only detects these sensor extrinsic errors when they result in the IOU of a track dropping below the 50\% threshold, which causes there to no longer be a match to the rest of the CAV/CIS report and finally the offending CAV/CIS is punished for not detecting the object entirely. Conversely, \conclave{} has a concept for how much variance it expects in sensing error and as that variance starts to leave the expected limits, the vehicle trust score is punished for that. This variance threshold results in quick detection of minor extrinsics errors which are missed by TruPercept. RMSE of Conlcave in all tests for E1-E3 is 6.3\% higher than TruPercept and 10.3\% better than without trust scoring. 

\noindent{\textbf{5.2 - \conclave{} detects malicious errors faster than TruPercept}}
E4-E6 consist of accidental track removal as well as malicious injection and removal, seen in figure \ref{fig:e456_mttd}. These tests are similar to what TruPercept was designed for, with the caveat that we only have a single vehicle experiencing the error, whereas TruPercept had all vehicles experiencing the same probability of error \cite{hurl2020trupercept}. TruPercept performs well in this case, detecting 34\% of 300 tests. \conclave{} beats TruPercept, detecting 67.3\% of the 300 tests. This is because a malicious actor that is injecting fake vehicles into their data will not purposely report a low camera confidence. Therefore, in the second test, TruPercept relies completely on the IOU mismatch of detections to identify a problem where \conclave{} can detect higher variances and report errors sooner. For the RMSE case, we can see that both \conclave{} and TruPercept respond to E4, so \conclave{} is only 1.9\% better than TruPercept and 5. 1\% better than without trust scoring. Although these error injections have a high probability, they are filtered out by local sensor fusion, so the RMSE effect is less than E1-E3.

\noindent{\textbf{5.3 - \conclave{} detects more software errors, and faster}}
\newline E7-E9 look at common cases of mis-weighting. Overall Trupercept detected 32.6\% of the 300 tests while \conclave{} was able to detect 83.5\%, which can be seen in figure \ref{fig:e789_mttd}. For the E7 localization error and E8 local sensor fusion misweight, we can see that \conclave{} detects it but Trupercept misses it due to \conclave{} being tuned to detect variance while Trupercpet is not. E9 on the other hand is detected by both methods with \conclave{} just slightly edging ahead in detection speed. Again, this is due to the nature of \conclave{} being able to detect larger and smaller variance than expected and report the errors quickly. Meanwhile TruPercept has to wait until detections start to mismatch IOU wise before it will start to detect the errors. Meanwhile, TruPercept takes longer to respond and therefore has a worse result. RMSE of \conclave{} was 3.9\% better than Trupercept and 5.8\% better than no trust scoring.

\noindent{\textbf{5.4 - \conclave{} detects many communication faults and attacks}}
E10 - E14 showcase tolerance to a variety of errors and attack vectors that are not typically captured by a trust scoring system alone. This is apparent when looking at figure \ref{fig:eothers_mttd} on the bottom right where only one of the errors, E14, is detected by Trupercept. Trupercept detected 19.6\% of communication faults and attacks while \conclave{} detected a perfect 100\% of the 200 tests. Furthermore, \conclave{} detected E10-13 in less than two seconds, or two consensus rounds. We did not compare RMSE because Trupercept as well as the performance of the baseline technique, was rendered inoperable in the case of E10, E11, and E13 beyond recovery.

% \subsubsection{\textbf{\conclave{} can run in real-time on embedded hardware}}
% We tested \conclave{} onboard our 1/10 scale vehicle hardware in a typical case and found that Authentication and Consensus round took on average 0.052s for 2 CAVs and 1 CIS, and 0.074s for the case of 4 CAVs and 2 CISs. This was running on a Jetson Nano on a Quad-core ARM A57 @ 1.43GHz processor using a WN725N USB network adapter and a TP Link AC1200 router as the communication backbone. Playing the part of the RSU was an Intel core i7 3770 which clocked an average time for cooperative perception and trust scoring of 0.12s for the 2 CAVs and 1 CIS case and 0.18s for the 4 CAVs and 2 CIS. All code was written in Python and we made no attempts to optimize it so we expect this would be much faster on modern hardware using more optimized code.

\vspace{-5pt}
\section{Conclusion}
In this paper we present a method to secure cooperatively perception-based applications for connected autonomous vehicles that we call \conclave{}. \conclave{} consists of three parts, an authentication method, a consensus round, and a trust scoring method that are pipelined such that it can be run in real time. \conclave{} was able to detect more categories of faults and errors, including both malicious and unintentional errors, while being faster than the state of the art method TruPercept. In future work, we would like to expand \conclave{} to work for all cooperative driving scenarios, including those that need path plan trust scoring. 

\vspace{-5pt}
\bibliographystyle{ACM-Reference-Format}
\bibliography{conclave}


\begin{thebibliography}{22}

%%% the Web address.

\ifx \showCODEN    \undefined \def \showCODEN     #1{\unskip}     \fi
\ifx \showDOI      \undefined \def \showDOI       #1{#1}\fi
\ifx \showISBNx    \undefined \def \showISBNx     #1{\unskip}     \fi
\ifx \showISBNxiii \undefined \def \showISBNxiii  #1{\unskip}     \fi
\ifx \showISSN     \undefined \def \showISSN      #1{\unskip}     \fi
\ifx \showLCCN     \undefined \def \showLCCN      #1{\unskip}     \fi
\ifx \shownote     \undefined \def \shownote      #1{#1}          \fi
\ifx \showarticletitle \undefined \def \showarticletitle #1{#1}   \fi
\ifx \showURL      \undefined \def \showURL       {\relax}        \fi
% invisible to TeX
\providecommand\bibfield[2]{#2}
\providecommand\bibinfo[2]{#2}
\providecommand\natexlab[1]{#1}
\providecommand\showeprint[2][]{arXiv:#2}

\bibitem[Andert et~al\mbox{.}(2022)]%
        {andert2022paramertized}
\bibfield{author}{\bibinfo{person}{Edward Andert} {et~al\mbox{.}}} \bibinfo{year}{2022}\natexlab{}.
\newblock \showarticletitle{Accurate Cooperative Sensor Fusion by Parameterized Covariance Generation for Sensing and Localization Pipelines in CAVs}. In \bibinfo{booktitle}{\emph{2022 IEEE 25th International Conference on Intelligent Transportation Systems (ITSC)}}. IEEE, \bibinfo{pages}{1--8}.
\newblock


\bibitem[Behrens et~al\mbox{.}(2020)]%
        {behrens2020pando}
\bibfield{author}{\bibinfo{person}{Hans~Walter Behrens} {et~al\mbox{.}}} \bibinfo{year}{2020}\natexlab{}.
\newblock \showarticletitle{Pando: Efficient Byzantine-Tolerant Distributed Sensor Fusion using Forest Ensembles}. In \bibinfo{booktitle}{\emph{ICC 2020-2020 IEEE International Conference on Communications (ICC)}}. IEEE, \bibinfo{pages}{1--6}.
\newblock


\bibitem[Bellare et~al\mbox{.}(1997)]%
        {bellare1997new}
\bibfield{author}{\bibinfo{person}{Mihir Bellare} {et~al\mbox{.}}} \bibinfo{year}{1997}\natexlab{}.
\newblock \showarticletitle{A new paradigm for collision-free hashing: Incrementality at reduced cost}. In \bibinfo{booktitle}{\emph{International Conference on the Theory and Applications of Cryptographic Techniques}}. Springer, \bibinfo{pages}{163--192}.
\newblock


\bibitem[Bhattacharya et~al\mbox{.}(2022)]%
        {bhattacharya20226blocks}
\bibfield{author}{\bibinfo{person}{Pronaya Bhattacharya} {et~al\mbox{.}}} \bibinfo{year}{2022}\natexlab{}.
\newblock \showarticletitle{6G-enabled trust management scheme for decentralized autonomous vehicles}.
\newblock \bibinfo{journal}{\emph{Computer Communications}}  \bibinfo{volume}{191} (\bibinfo{year}{2022}), \bibinfo{pages}{53--68}.
\newblock


\bibitem[Cavorsi et~al\mbox{.}(2022)]%
        {cavorsi2022exploiting}
\bibfield{author}{\bibinfo{person}{Matthew Cavorsi} {et~al\mbox{.}}} \bibinfo{year}{2022}\natexlab{}.
\newblock \showarticletitle{Exploiting trust for resilient hypothesis testing with malicious robots}.
\newblock \bibinfo{journal}{\emph{arXiv preprint arXiv:2209.12285}} (\bibinfo{year}{2022}).
\newblock


\bibitem[Dharminder et~al\mbox{.}(2021)]%
        {dharminder2021edge}
\bibfield{author}{\bibinfo{person}{Dharminder Dharminder} {et~al\mbox{.}}} \bibinfo{year}{2021}\natexlab{}.
\newblock \showarticletitle{Edge based authentication protocol for vehicular communications without trusted party communication}.
\newblock \bibinfo{journal}{\emph{Journal of Systems Architecture}}  \bibinfo{volume}{119} (\bibinfo{year}{2021}), \bibinfo{pages}{102242}.
\newblock


\bibitem[Dibaei et~al\mbox{.}(2020)]%
        {dibaei2020attacks}
\bibfield{author}{\bibinfo{person}{Mahdi Dibaei} {et~al\mbox{.}}} \bibinfo{year}{2020}\natexlab{}.
\newblock \showarticletitle{Attacks and defences on intelligent connected vehicles: A survey}.
\newblock \bibinfo{journal}{\emph{Digital Communications and Networks}} \bibinfo{volume}{6}, \bibinfo{number}{4} (\bibinfo{year}{2020}), \bibinfo{pages}{399--421}.
\newblock


\bibitem[Guo et~al\mbox{.}(2020)]%
        {guo2020proof}
\bibfield{author}{\bibinfo{person}{Hao Guo} {et~al\mbox{.}}} \bibinfo{year}{2020}\natexlab{}.
\newblock \showarticletitle{Proof-of-event recording system for autonomous vehicles: A blockchain-based solution}.
\newblock \bibinfo{journal}{\emph{IEEE Access}}  \bibinfo{volume}{8} (\bibinfo{year}{2020}), \bibinfo{pages}{182776--182786}.
\newblock


\bibitem[Halderman et~al\mbox{.}(2009)]%
        {halderman2009lest}
\bibfield{author}{\bibinfo{person}{J.~Alex Halderman} {et~al\mbox{.}}} \bibinfo{year}{2009}\natexlab{}.
\newblock \showarticletitle{Lest We Remember: Cold-Boot Attacks on Encryption Keys}.
\newblock \bibinfo{journal}{\emph{Commun. ACM}} \bibinfo{volume}{52}, \bibinfo{number}{5} (\bibinfo{date}{May} \bibinfo{year}{2009}), \bibinfo{pages}{91--98}.
\newblock
\showISSN{0001-0782, 1557-7317}


\bibitem[Han et~al\mbox{.}(2022)]%
        {han2022distributed}
\bibfield{author}{\bibinfo{person}{Jinheng Han} {et~al\mbox{.}}} \bibinfo{year}{2022}\natexlab{}.
\newblock \showarticletitle{Distributed finite-time safety consensus control of vehicle platoon with senor and actuator failures}.
\newblock \bibinfo{journal}{\emph{IEEE Transactions on Vehicular Technology}} \bibinfo{volume}{72}, \bibinfo{number}{1} (\bibinfo{year}{2022}), \bibinfo{pages}{162--175}.
\newblock


\bibitem[Hbaieb et~al\mbox{.}(2022)]%
        {hbaieb2022survey}
\bibfield{author}{\bibinfo{person}{Amal Hbaieb} {et~al\mbox{.}}} \bibinfo{year}{2022}\natexlab{}.
\newblock \showarticletitle{A survey of trust management in the Internet of Vehicles}.
\newblock \bibinfo{journal}{\emph{Computer Networks}}  \bibinfo{volume}{203} (\bibinfo{year}{2022}), \bibinfo{pages}{108558}.
\newblock


\bibitem[He et~al\mbox{.}(2019)]%
        {he2019cooperative}
\bibfield{author}{\bibinfo{person}{Jianhua He} {et~al\mbox{.}}} \bibinfo{year}{2019}\natexlab{}.
\newblock \showarticletitle{Cooperative connected autonomous vehicles (CAV): research, applications and challenges}. In \bibinfo{booktitle}{\emph{2019 IEEE 27th International Conference on Network Protocols (ICNP)}}. IEEE, \bibinfo{pages}{1--6}.
\newblock


\bibitem[Hurl et~al\mbox{.}(2020)]%
        {hurl2020trupercept}
\bibfield{author}{\bibinfo{person}{Braden Hurl} {et~al\mbox{.}}} \bibinfo{year}{2020}\natexlab{}.
\newblock \showarticletitle{Trupercept: Trust modelling for autonomous vehicle cooperative perception from synthetic data}. In \bibinfo{booktitle}{\emph{2020 IEEE Intelligent Vehicles Symposium (IV)}}. IEEE, \bibinfo{pages}{341--347}.
\newblock


\bibitem[Khayatian et~al\mbox{.}(2020)]%
        {khayatian2020survey}
\bibfield{author}{\bibinfo{person}{Mohammad Khayatian} {et~al\mbox{.}}} \bibinfo{year}{2020}\natexlab{}.
\newblock \showarticletitle{A survey on intersection management of connected autonomous vehicles}.
\newblock \bibinfo{journal}{\emph{ACM Transactions on Cyber-Physical Systems}} \bibinfo{volume}{4}, \bibinfo{number}{4} (\bibinfo{year}{2020}), \bibinfo{pages}{1--27}.
\newblock


\bibitem[Liu et~al\mbox{.}(2022)]%
        {liu2022frequency}
\bibfield{author}{\bibinfo{person}{Chen Liu} {et~al\mbox{.}}} \bibinfo{year}{2022}\natexlab{}.
\newblock \showarticletitle{Frequency {{Throttling Side-Channel Attack}}}. In \bibinfo{booktitle}{\emph{Proceedings of the 2022 {{ACM SIGSAC Conference}} on {{Computer}} and {{Communications Security}}}} \emph{(\bibinfo{series}{{{CCS}} '22})}. \bibinfo{publisher}{{Association for Computing Machinery}}, \bibinfo{address}{{New York, NY, USA}}, \bibinfo{pages}{1977--1991}.
\newblock
\showISBNx{978-1-4503-9450-5}


\bibitem[Lu et~al\mbox{.}(2018)]%
        {lu2018survey}
\bibfield{author}{\bibinfo{person}{Zhaojun Lu} {et~al\mbox{.}}} \bibinfo{year}{2018}\natexlab{}.
\newblock \showarticletitle{A survey on recent advances in vehicular network security, trust, and privacy}.
\newblock \bibinfo{journal}{\emph{IEEE Transactions on Intelligent Transportation Systems}} \bibinfo{volume}{20}, \bibinfo{number}{2} (\bibinfo{year}{2018}), \bibinfo{pages}{760--776}.
\newblock


\bibitem[Mankodiya et~al\mbox{.}(2021)]%
        {mankodiya2021xai}
\bibfield{author}{\bibinfo{person}{Harsh Mankodiya} {et~al\mbox{.}}} \bibinfo{year}{2021}\natexlab{}.
\newblock \showarticletitle{XAI-AV: Explainable artificial intelligence for trust management in autonomous vehicles}. In \bibinfo{booktitle}{\emph{2021 International Conference on Communications, Computing, Cybersecurity, and Informatics (CCCI)}}. IEEE, \bibinfo{pages}{1--5}.
\newblock


\bibitem[Nandy et~al\mbox{.}(2021)]%
        {nandy2021secure}
\bibfield{author}{\bibinfo{person}{Tarak Nandy} {et~al\mbox{.}}} \bibinfo{year}{2021}\natexlab{}.
\newblock \showarticletitle{A {{Secure}}, {{Privacy-Preserving}}, and {{Lightweight Authentication Scheme}} for {{VANETs}}}.
\newblock \bibinfo{journal}{\emph{IEEE Sensors Journal}} (\bibinfo{year}{2021}), \bibinfo{pages}{1--1}.
\newblock
\showISSN{1558-1748}


\bibitem[Sabahi(2011)]%
        {sabahi2011security}
\bibfield{author}{\bibinfo{person}{Farzad Sabahi}.} \bibinfo{year}{2011}\natexlab{}.
\newblock \showarticletitle{The security of vehicular adhoc networks}. In \bibinfo{booktitle}{\emph{2011 third international conference on computational intelligence, communication systems and networks}}. IEEE, \bibinfo{pages}{338--342}.
\newblock


\bibitem[Song et~al\mbox{.}(2008)]%
        {song2008bosco}
\bibfield{author}{\bibinfo{person}{Yee~Jiun Song} {et~al\mbox{.}}} \bibinfo{year}{2008}\natexlab{}.
\newblock \showarticletitle{Bosco: One-step byzantine asynchronous consensus}. In \bibinfo{booktitle}{\emph{Distributed Computing: 22nd International Symposium, DISC 2008, Arcachon, France, September 22-24, 2008. Proceedings 22}}. Springer, \bibinfo{pages}{438--450}.
\newblock


\bibitem[Svensson et~al\mbox{.}(2011)]%
        {svensson2011set}
\bibfield{author}{\bibinfo{person}{Lennart Svensson} {et~al\mbox{.}}} \bibinfo{year}{2011}\natexlab{}.
\newblock \showarticletitle{Set JPDA filter for multitarget tracking}.
\newblock \bibinfo{journal}{\emph{IEEE Transactions on Signal Processing}} \bibinfo{volume}{59}, \bibinfo{number}{10} (\bibinfo{year}{2011}), \bibinfo{pages}{4677--4691}.
\newblock


\bibitem[Wang et~al\mbox{.}(2021)]%
        {wang2021blockchain}
\bibfield{author}{\bibinfo{person}{Weizheng Wang} {et~al\mbox{.}}} \bibinfo{year}{2021}\natexlab{}.
\newblock \showarticletitle{Blockchain and {{PUF-based Lightweight Authentication Protocol}} for {{Wireless Medical Sensor Networks}}}.
\newblock \bibinfo{journal}{\emph{IEEE Internet of Things Journal}} (\bibinfo{year}{2021}), \bibinfo{pages}{1--1}.
\newblock
\showISSN{2327-4662}


\end{thebibliography}

\end{document}